\pdfoutput=1

\documentclass[11pt]{article}

\usepackage[]{acl}

\usepackage{times}
\usepackage{latexsym}

\usepackage[T1]{fontenc}

\usepackage[utf8]{inputenc}

\usepackage{microtype}

\usepackage{graphicx}
\usepackage{svg}
\usepackage{amsmath}

\usepackage{kotex}
\usepackage{enumitem}
\usepackage{url}
\usepackage{multirow}
\usepackage{hyperref}[breaklinks,colorlinks]

\usepackage{microtype}

\title{APEACH: Attacking Pejorative Expressions 
with Analysis on Crowd-Generated Hate Speech Evaluation Datasets
}

\author{Kichang Yang$^{1,2,3}$\thanks{\;\;Equal contribution.} \thanks{\;\;Current Affiliation: LG AI Research} \thanks{\;\;Contact: ykcha9@gmail.com},  Wonjun Jang$^{1,3}$\footnotemark[1] ,  Won Ik Cho$^{4}$\footnotemark[1] \\
        $^1$Kakao Corp.\\
        $^2$Kakao Enterprise Corp.\\
        $^3$School of Software, Soongsil University \\
        $^4$Dept. of ECE, Seoul National University
        \\\texttt{\{kevin.ai,joel.j\}@kakaocorp.com} 
        \\\texttt{tsatsuki@snu.ac.kr}\\
        }

\date{}

\begin{document}
\maketitle
\begin{abstract}
In hate speech detection, developing training and evaluation datasets across various domains is the critical issue. Whereas, major approaches crawl social media texts and hire crowd-workers to annotate the data. Following this convention often restricts the scope of pejorative expressions to a single domain lacking generalization. Sometimes domain overlap between training corpus and evaluation set overestimate the prediction performance when pretraining language models on low-data language. To alleviate these problems in Korean, we propose APEACH that asks unspecified users to generate hate speech examples followed by minimal post-labeling. We find that APEACH can collect useful datasets that are less sensitive to the lexical overlaps between the pretraining corpus and the evaluation set, thereby properly measuring the model performance.

\end{abstract}
\begin{figure*}
   \centering
   \includegraphics[scale=0.85]{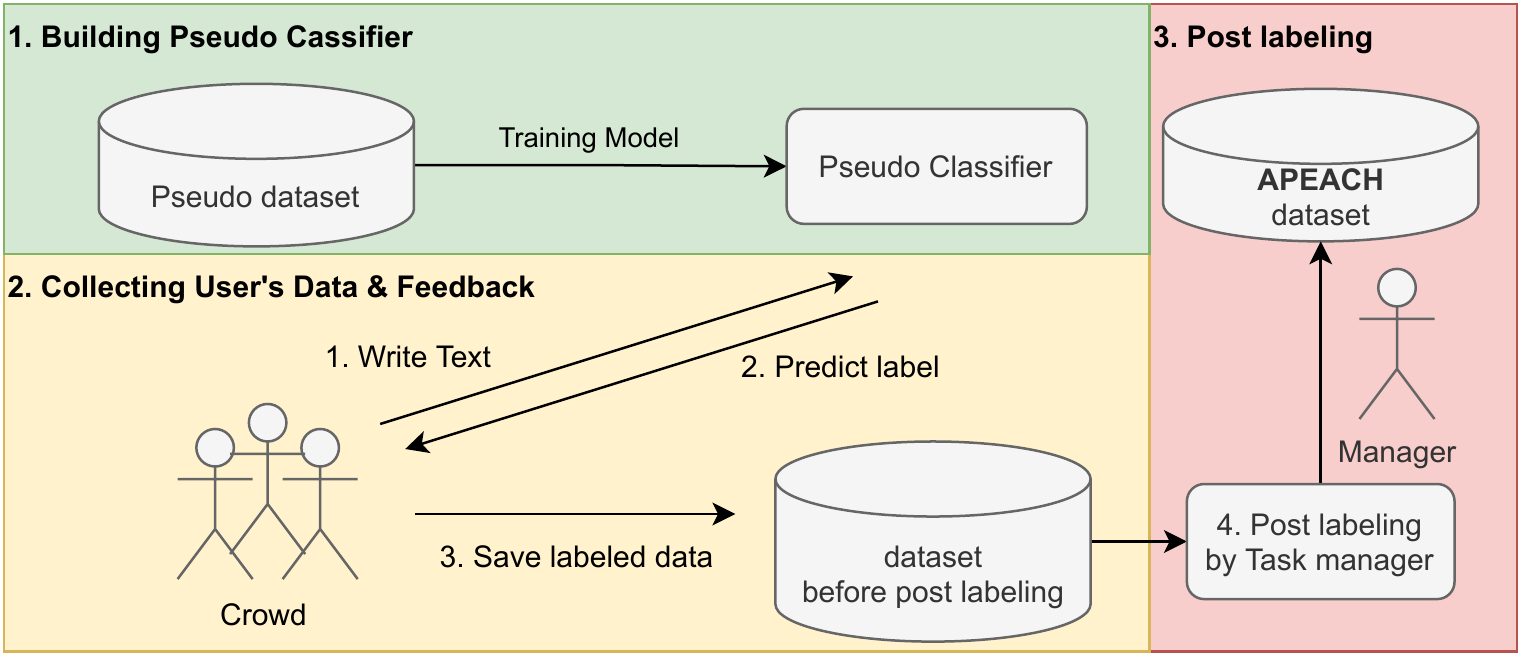}
   \caption{Overall schematic process of the proposed system (APEACH). Through this process, we can create datasets that are less vulnerable to probable bias in domain and style of the text. Also, creating datasets using this scheme can help prevent license and privacy issues.  
   } \label{fig:figure1}
\end{figure*} 

\section{Introduction}
Detecting toxic or pejorative expressions has been a crucial issue in various online communities. In particular, flaming or trolling in online communities is regarded as hostile behavior that can disrupt the public order and cause mental harm to individuals and groups. Attempts to define and detect hate speech from a natural language processing (NLP) perspective have called for timely works, and notable approaches have been suggested so far. In specific, \citet{waseem2016hateful} primarily attacked the judgment on hate speech for Twitter text, and  \citet{hateoffensive} further investigated the offensiveness of social media texts beyond binary detection. Recently, \citet{huang-etal-2020-multilingual} have suggested how demographic matters in hate speech analysis, for the corpora of five different languages, discerning the multilingual tendency.

Creating a hate speech\footnote{Though toxic or pejorative expressions are distinguished with `hate speech', a more political term, here we interchangeably use both sides of terms for our description and alignment with previous works.} dataset generally involves annotating short documents such as web text, and the context where the hate expressions are from may or may not be given in the process. However, annotating on existing web text has several limitations that deter the dataset's reliability. 
First, the corpus may incorporate the potential risk of license and personally identifiable information issues that come from the characteristics of online materials. Next, if the text is crawled from a restricted scope of domain, the topics of examples may not be diverse, which can result in the evaluation that focuses only a part of social issues (e.g., gender). Moreover, in view of model training, using a specific domain of web text that might have been used as a pretraining corpus of public language models, may intervene the fair competition between models and mislead the evaluation.
Such tendency is more apparent in non-English hate speech detection tasks where the training and evaluation of language models rely upon only a small number of benchmarks. For instance, in Korean, the usual performance checking of pretrained language models (PLMs) adopts BEEP! \cite{moon-etal-2020-beep}, a currently available hand-labeled hate speech dataset in Korean where the domain of raw text is celebrity news comment. People found out that the PLMs trained upon news comments perform better than others which base on news or Wikipedia, but it throws a question to the generalizability and fairness of the evaluation.

How can we address such limitation of annotation-based web text corpus construction? Though collecting a variety of data as in multi-genre natural language inference (MultiNLI) \cite{williams2018broad} might be the most intuitive solution, it requires  economic and human resources and still does not resolve the issue of corpus overlap. In this regard, we hypothesized it a reasonable approach to let anonymous paid workers generate toxic expressions, where the expressions are generated from scratch with a minimum guideline. 

At a glance, simply opening a web page for text collecting and encouraging user participation seemed to work. However, we noted that neither the data quality nor the open license of the output could be guaranteed in those processes. Thus, we established a crowd-driven hate speech generation scheme using a moderator to ensure the privacy of hate speech authors and obtain a quality-checked corpus at the same time. In specific, we adopt crowd-sourcing platform (as a moderator) and workers for the paid writing, provided with the prompts to guide the generation, to achieve diverse hate speech and prevent participants' disgrace. 
For the facilitation of text generation and collection process in our research, we devise `\textbf{System}', an environment that interacts with the \textbf{crowd} (who provides the data) and the \textbf{task manager} (who collects the data), which is composed of i) building a hate speech pseudo-classifier and deploying the model, ii) collecting user-generated data and feedback, and iii) post-labeling of the task managers (Figure~\ref{fig:figure1}).

Followingly, we obtain \textbf{APEACH}, which denotes the collecting scheme and the resulting dataset at the same time. It contains about 3K instances for Korean hate speech detection evaluation; the corpus is well balanced in sentence length and topics, also being aligned with the models trained with the existing hate speech dataset (BEEP!). Most of all, by comparing the model performances of our dataset using publicly available PLMs where the pretraining corpus overlap with BEEP!, we prove that APEACH is less vulnerable to misleading results that might come from the corpus-level similarity with the pretraining corpus.
Our contribution to this field is as follows:

\begin{itemize}
\setlength\itemsep{.2em}
    \item Propose a scheme that collects user-generated hate speech from scratch without undertaking conventional annotation process.
    \item Build and release a new evaluation set for Korean hate speech detection,\footnote{\url{https://huggingface.co/datasets/jason9693/APEACH}} free from license and privacy issues, preventing potential bias of corpus domains. 
    \item Conduct a model-based comparison with another human-annotated hate speech benchmark, showing that the generalizability of the proposed evaluation set is implied from less overlap with specific pretraining corpus.
\end{itemize}


\section{System}
\label{sec:system}

Our system consists of three processes: 1) building pseudo-classifier, 2) collecting users' data \& feedback, and 3) post-labeling. First two are to be described in this section.

\begin{figure} [h]
   \centering
  \includegraphics[width=\columnwidth]{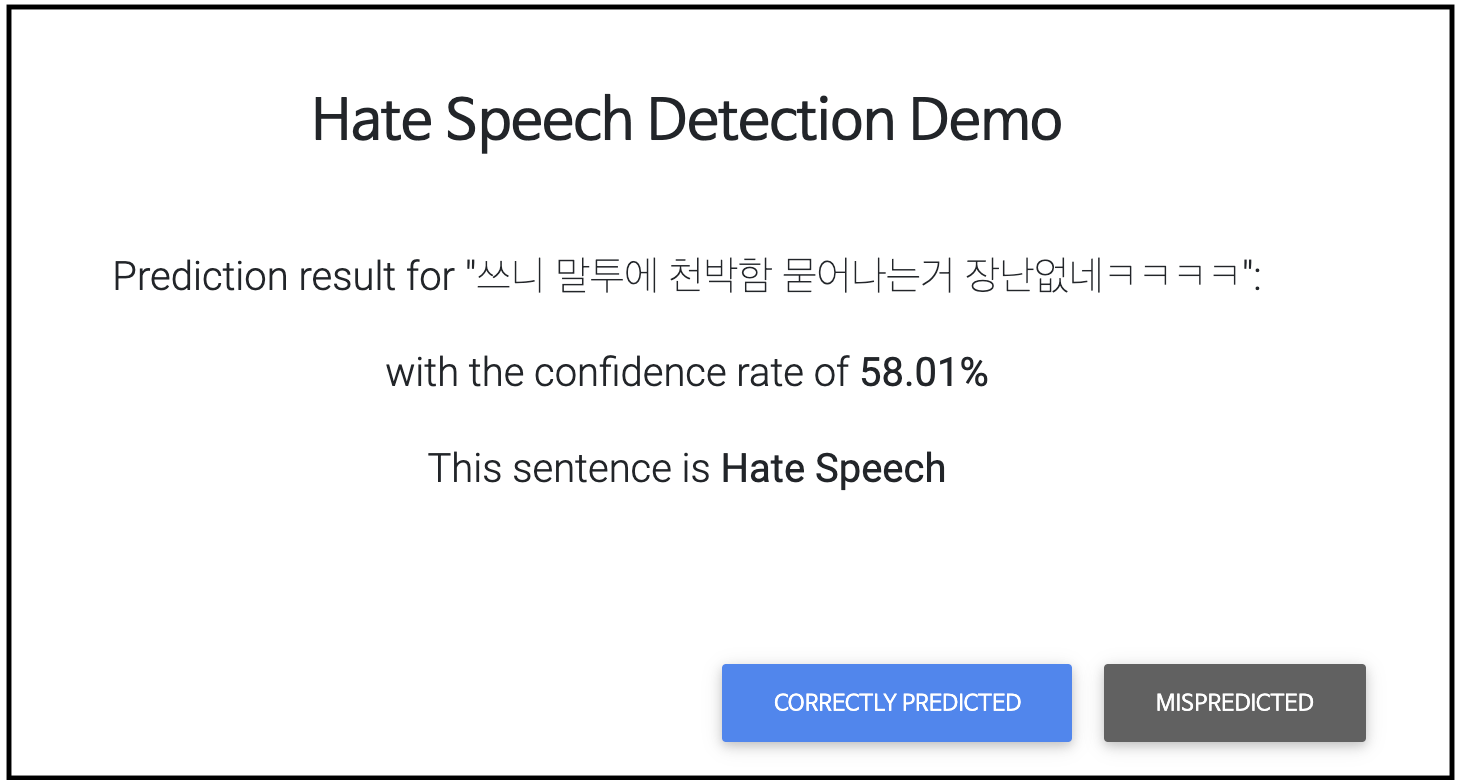}
   \caption{
   A screen in which users enter input sentences to the deployed model and check the predicted result. At the bottom right is a screen where they can select whether the predicted result is correct (blue) or not (grey). The translation for the input sentence is: ``\textit{what an uneducated lol }''.} \label{fig:figure2}
\end{figure} 
\subsection{Building a Pseudo-Classifier}
To build the system from scratch, we deploy a pseudo-classifier to compensate for the crowd's loss of concentration coming from repetitive tasks and let them participate actively in the collection process. For this, we primarily created a pseudo-labeled dataset to train the classifier, not for evaluation. As usual hate speech detectors, the pseudo-classifier receives the user-generated text as an input and predicts whether the text incorporates bias or toxicity. However, since this is not for real service but a tool for  participants'  confirmation on the label of their work, we trained the classifier that displays only a basic performance. 
We simply create a dictionary of profanity terms to obtain a pseudo-labeled web text dataset, and use it to train a simple binary classifier. The details of dataset construction and model selection are provided in Appendix~\ref{sec:supplementary}.

\subsection{Deployment and Text Collection}
The pseudo-classifier is deployed to a server to collect user input and feedback. As in Figure~\ref{fig:figure1}, when the user enters the test input, the predicted label is displayed. The user further determines whether the corresponding label equals the user's original intention, namely whether it is hate speech or not. The user interface (UI) for the prediction and feedback is exhibited in Figure~\ref{fig:figure2}.

In specific, in the user feedback phase, two buttons are provided, namely ``correctly predicted'' and ``mispredicted'' (Figure~\ref{fig:figure2}). If ``correctly predicted'', the prediction is saved along with the user input as the ground truth. In the case of ``mispredicted'', the ground truth is saved as a reversed version of the prediction.  

\section{Dataset}
\label{sec:dataset}
Using the above system, we construct an evaluation dataset with the user-generated data that is inspected with a model prediction.

\subsection{Prompts for Text Generation}
In general, hate speech includes flamings observed in the web communities, namely the threatening expressions that are represented as text or even in some multi-modal format \cite{NEURIPS2020_1b84c4ce}. They can be expressed in hostile or discriminating words. However, letting the participants merely generate the hate speech from scratch might be challenging and misleading.

\begin{figure}
   \centering
   \includegraphics[width=\columnwidth]{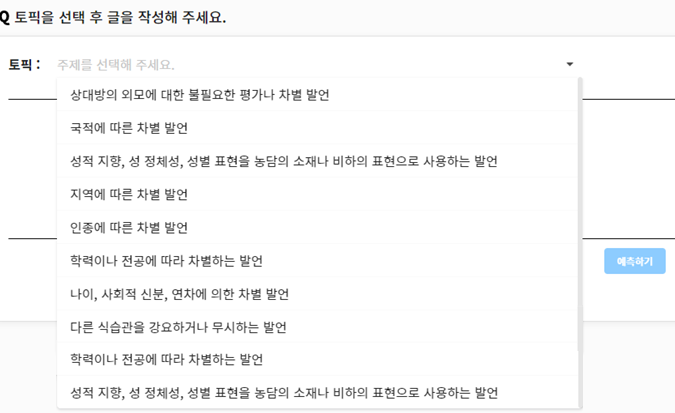}
   \caption{Web interface utilized in the crowd-sourcing process. } \label{fig:figure4}
\end{figure}

\paragraph{Topic} For effective and efficient data collection, we provide the participants with criteria on various topics of hate speech. 
We set ten topics that the participants can refer to in generating the text, which is inspired by the code of conduct (COC) of PyCon KR. \footnote{\url{https://www.pycon.kr/en/2020/about/coc/}} Each of the topics denotes the main attribute of the hate speech that is to be generated. 
\begin{enumerate}[noitemsep]
\item Behaviors based on gender stereotypes
\item Discrimination or demeaning jokes with one's sexual orientation or identity 
\item Discrimination or stereotypes on age, social status, or experience
\item Discrimination based on nationality/ethnicity
\item Racial discrimination
\item Discrimination based on origin or residence
\item Unnecessary or offensive judgments on one's appearance
\item Demeaning or offensive words for illness or disability
\item Forcing or depreciating with eating habits
\item Rude or discriminatory remarks based on others' academic background or major
\end{enumerate}
As shown in Figure~\ref{fig:figure4}, workers generate text by selecting one of the topics above. Prompts regarding the topic of hate speech are provided in a dropdown format, while the order of topics is randomly shuffled for each input. Workers enter the input after selecting one of them.


\paragraph{Label} We define whether a sentence contains hate expression or not as ``label''.  Besides the topic, the workers are assigned with either they should generate a sentence that contains hate speech or not. The latter case denotes a neutral or seemingly controversial utterance that only shares the topic with hate speeches but is not offensive; e.g., ``\textit{I hate those who demeans BLM movements.}'' for the topic `Racial discrimination'. In our study, the hate speech (positive sample) and non-hate speech (negative sample) serve as an element of the balanced dataset for the detection task.

\subsection{Post Labeling}
\label{sec:post_labeling}

For each user input, the pseudo-classifier yields the prediction. For the assigned label, the user input is confirmed based on the user's feedback. For instance, when the assigned label is hate speech, and if the model yields `non-hate speech', the user may check ``mispredicted'' and the ground truth is saved as ``hate speech''. In this process, if the assigned label differs from the saved ground truth, we conclude that the user mislabeled or misunderstood the label, and  automatically remove the instance.
\footnote{ 
Note that this process is a simple but strong checkpoint for the robustness of the labels in the dataset, leveraging the possible human error of the user feedback system \cite{stumpf2007toward}. Given that the participants are aware of the pseudo-classification, the system allows them to deem the toxicity of the contents they generate.
}

We faced some questionable instances that came from the diverse ethical standards of the participants. However, to guarantee the characteristics of the crowd-generated hate speech dataset, such erroneous cases were checked with the minor engagement of task managers. We call this process \textit{post-labeling}, which ensures the quality of the dataset by applying a conventional annotation and voting process in the final decision. In specific, three task managers, who are speakers of contemporary Korean, checked if the user's feedback was appropriate for each instance. Since we regard the user's choice as ground truth, we only dropped the instances that all of the three task managers found irrelevant with the assigned label. 

\subsection{Dataset Collection}
As discussed, the most intuitive way for construction is to collect text inputs from an unspecified crowd using an online platform, e.g., a demo page. Hate speech detector is itself a great contribution to the community, thus getting user feedback with a closed or open beta service is not an unnatural choice, in view of both research progress and industrial development. 

However, such an approach incorporates some critical issues on data quality and privacy, which originates in the characteristics of hate speech text. It is widely known that the reliability of a generated corpus is not usually guaranteed if there is a lack of time or budget that it takes in creating the dataset. We guessed that people might not reveal their identity to receive compensation for their toxic expressions, mainly due to the fear of being recognized as a politically incorrect person. To confirm that our guess is correct, a web-based pilot resulted in the collection of text with degraded quality, sometimes violating the license issues or containing personally identifiable information. \footnote{We want to point out that the pilot only provided us with motivation, and was not a part of our experiment. The pilot demo was organized in the very first phase of our project to check how our pseudo-classifier deals with real-world toxic expressions, on a publicly accessible web site, to found out that undisclosed crowds show uncontrollable behaviors. We became aware of legal and ethical standards that are required in hate speech dataset creation, and did not collect or store any logs or data of user input queries (nor they were used in our experiments or analyses). Our final dataset is based on the manual generation of paid workers, not a web-based collection.} Also, such an approach might not be approved by research communities and institutional review board (IRB). 
In order to cope with the limitation of unspecified user-generated hate speech collection discussed above, we create a dataset leveraging the worker pool of the crowd-sourcing platform, while taking the same user generation guideline and post-labeling scheme.

\paragraph{Compensation through moderator}
Workers must be identified for the compensation of the hate speech generation, but it may harm the anonymity of the collection phase and eventually affect the natural text generation. Therefore, we enable the crowd-sourcing platform to play a role as a \textbf{moderator} between the task managers and workers. In other words, the project is designed in the way that only the moderator manages the workers' profiles, preventing them from being known to the task managers. In this way, we can accommodate both compensation and anonymity for the workers.

\paragraph{Worker selection for dataset quality}
One of the aims of the evaluation set we construct is to reflect the diversity of contents as much as possible. However, not all the paid workers of the crowd-sourcing platform are qualified for our project. Thus, we had a tutorial to prevent the low-quality generation which might take place in unconstrained user data collections. In detail, we receive ten inputs per worker and count the portion of mislabeled instances to drop the workers with frequent faults. Also, we checked the worker's sincerity with i) if the input is longer than a single character and ii) if the input does not replicate the examples in the guideline. 
154 out of 230 workers were finally admitted for participation in the main construction. 


\paragraph{Diversity of crowd-generated hate speech}
In contrast to the anonymous collection where the task managers find it difficult to ask the participants to sincerely generate various topics of texts owing to the lack of compensation, the proposed scheme helps attenuate such limitations with the utilization of topic prompt. It guides the text generation of the workers 
and helps collect non-hate speech that is less considered in the previous hate speech literature. In addition, to prevent the contents from being biased due to some heavy workers, we let the text generation be a maximum of 40 per participant. 
\subsection{Dataset Summary}

Our construction scheme can simultaneously guarantee data quality, topic variety/distribution, and ethical consideration of crowd-generated hate speech, with the crowd-sourcing platform as a moderator. Sentences are aggregated from the pilot and main collection phase. In the pilot, instances with clear disagreements were not selected for further project, and in the main phase, disagreements among task managers were re-labeled as an opposite class. We provide the detailed information on agreement in Appendix~\ref{sec:post-labeling}.

\begin{figure}
   \includegraphics[width=\columnwidth]{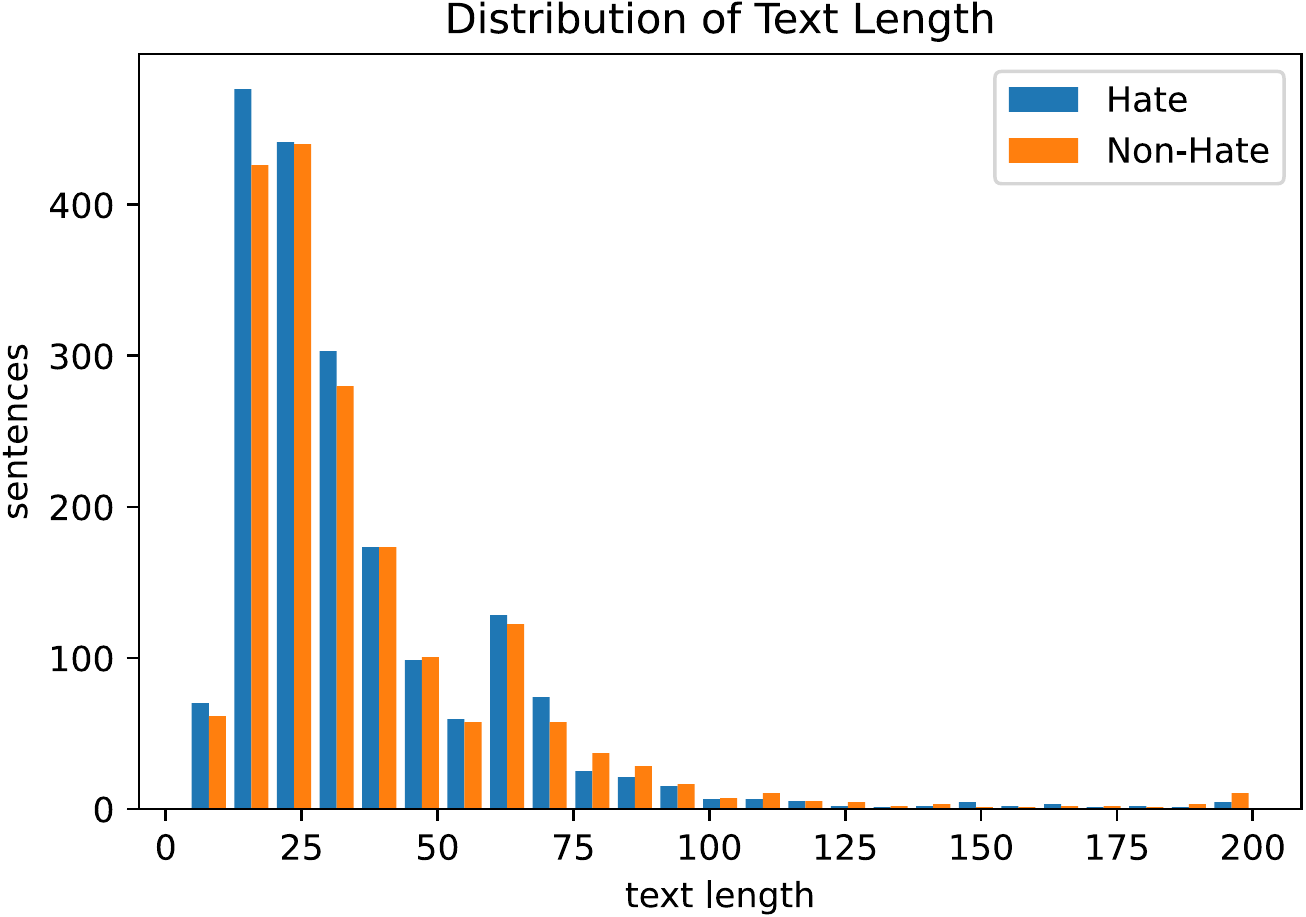}
   \caption{Distribution of text by length and label.} \label{fig:figure5}
\end{figure}

\begin{figure}[hbt!]
   \centering
   \includegraphics[width=\columnwidth]{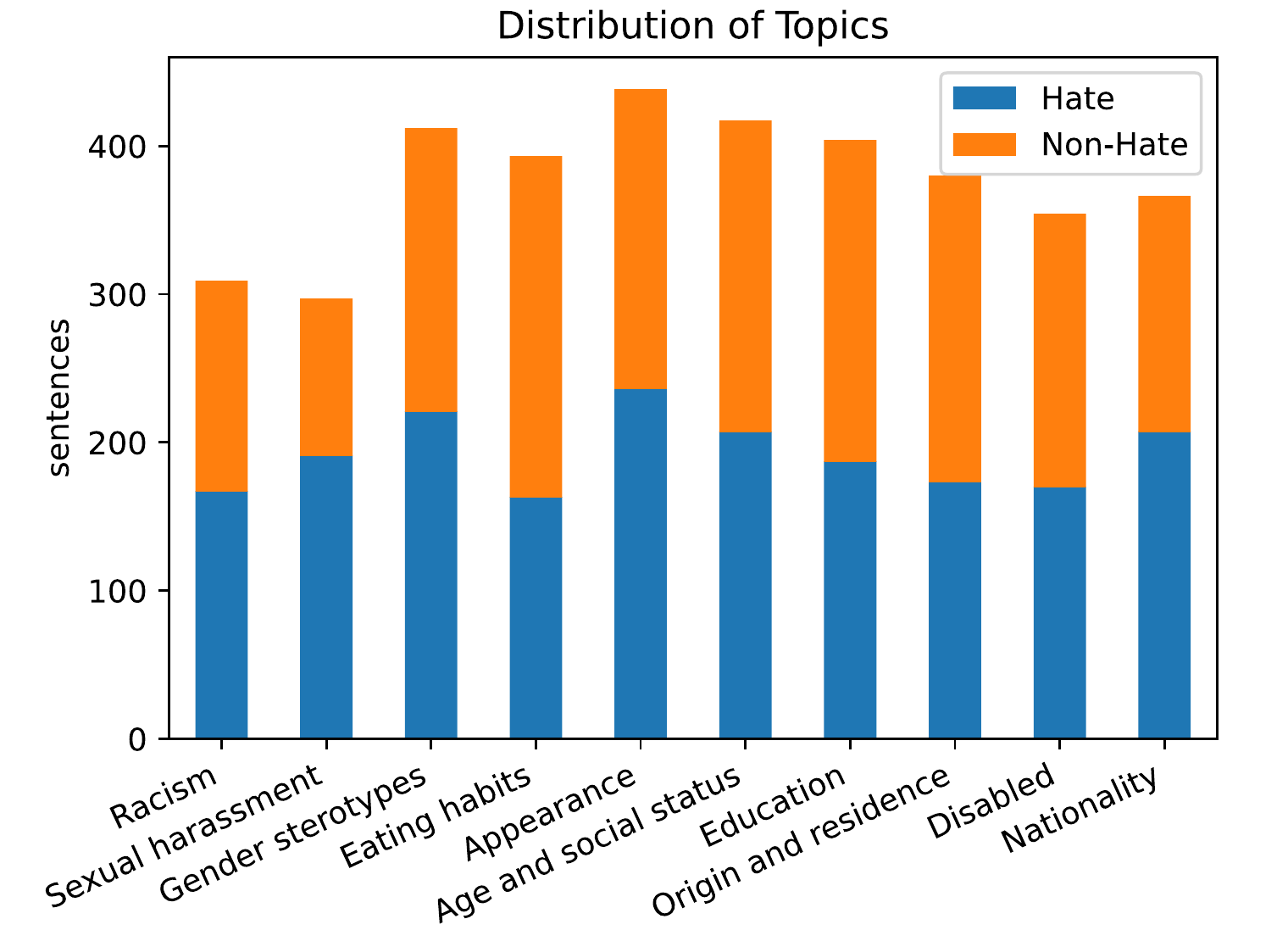}
   \caption{Topic/label distribution} \label{fig:figure8}
\end{figure}




\paragraph{Length distribution}
In Figure~\ref{fig:figure5},  similar length distribution is displayed between hate speech and non-hate speech in APEACH. This suggests that our construction scheme can prevent the biased distribution of the length of hate and non-hate speech. It enables us to further investigate if the hate speech detection model has handled the inductive bias coming from such distribution.





\paragraph{Distribution of topics}
By shuffling the order of topic prompts per every input, we prevent the bias which comes from the tendency that people habitually select the top candidate in the dropdown interface. 
In addition, we also confirmed that the two labels are evenly distributed in the dataset by assigning hate speech and non-hate speech in advance in the collection phase (Figure~\ref{fig:figure8}). 

\section{Experiment}
We exploit our corpus to evaluate the hate speech detection models trained with a widely used Korean hate speech benchmark, BEEP! \cite{moon-etal-2020-beep}. In specific, we compare APEACH (ours) and BEEP! dev set as an evaluation corpus, to check the generalizability and performance tendency using each set.


\begin{table*}
\centering
\resizebox{0.75\textwidth}{!}{%
\begin{tabular}{|c|c|c|c|}
\hline
\textbf{Model} & \textbf{BEEP! dev set} & \textbf{APEACH (ours)} & \textbf{Relative difference} \\
\hline\hline
KoBERT             & 0.8030     & 0.7885    & -1.81\%    \\
DistillKoBERT      & 0.7570       & 0.7715  & 1.92\%     \\
KoELECTRA-V3       & 0.7920       & 0.8101  & 2.29\%     \\
KcBERT-Base        & 0.8088      & 0.8086 & -0.02\%       \\
KcBERT-Large       & \textbf{0.8295}      & 0.8116 & -2.16\%       \\
SoongsilBERT-Base  & 0.8261      & \textbf{0.8424} & 1.97\%       \\
SoongsilBERT-Small & 0.8149      & 0.8228   & 0.97\%    \\
\hline\hline
\textbf{Composition}  
& \begin{tabular}{@{}c@{}} \small Hate + Offensive : 311 \\ \small None : 160 \\ \small Total : 471 \end{tabular} 
& \begin{tabular}{@{}c@{}} \small Hate : 1,922 \\ \small Non-hate : 1,848 \\ \small Total : 3770 \end{tabular} &  \\
\hline
\end{tabular}%
}
\caption{\label{result}
F1 score of binary classification performance for each model. The fine-tuning was conducted with the Beep! train set. At the bottom, we provide the number of sentences according to the labels of each dataset. 
}
\label{table1}
\end{table*}

\subsection{Korean Pretrained Language Models}
We adopt publicly available   Korean pretrained language models for the reproducibility. The characteristics of each model are as follows:
\begin{itemize}
\setlength\itemsep{.2em}
 \item \textbf{KoBERT \footnote{\url{https://github.com/SKTBrain/KoBERT}}, DistilKoBERT \footnote{\url{https://github.com/monologg/DistilKoBERT}}}: KoBERT is a PLM that follows the BERT \cite{devlin2019bert} training scheme, a self-supervised learning scheme based on Transformers \cite{vaswani2017attention} architecture, with Korean Wikipedia data.
 In addition, DistilKoBERT is a light-weighted version of KoBERT using distillation, adopting Huggingface's DistilBERT \cite{sanh2019distilbert} model.
\item \textbf{KoELECTRA} \footnote{\url{https://github.com/monologg/KoELECTRA}}: An ELECTRA \cite{clark2020electra}-based Korean language model pretrained with `Modu Corpus' released by the National Institute of Korean Language (NIKL) \cite{nikl2020corpora}, Korean Wikipedia, NamuWiki \footnote{A Large-scale Korean open encyclopedia. \url{https://namu.wiki/}}, and various news articles.
\item \textbf{KcBERT} \cite{lee2020kcbert} \footnote{\url{https://github.com/Beomi/KcBERT}}: A Korean BERT model trained with 12GB of NAVER politics news comments. \footnote{\url{https://bit.ly/3o1d7lk }}
 \item \textbf{SoongsilBERT} \footnote{\url{https://github.com/jason9693/Soongsil-BERT}}: In addition to the news comments data used by KcBERT, SoongsilBERT utilizes college community data and Modu Corpus. 
\end{itemize}
The architecture and fine-tuning configuration of each PLM are provided in Appendix~\ref{sec:training-detail}.

\subsection{Training Data}
We used BEEP! training set for fine-tuning the above PLMs. 
In detail, BEEP! is a human-annotated corpus where the intensity of hate speech is tagged with the labels of `hate', `offensive', and `none',  built upon celebrity news comments on a Korean online news platform. The instances with `hate' labels include hostile expressions, stigmatization, or sexual harassment, and `offensive' instances include sarcastic or inhumane expressions.

Although the construction scheme of train and dev set of BEEP! differs from our dataset, we want to compare the tendency between each set regarding hate speech detection models, utilizing both datasets. Nevertheless, since APEACH merely suffices the scale of an evaluation corpus, we first fine-tune the models with BEEP! train set.

\subsection{Evaluation}
We formulate BEEP! dev set and APEACH as both binary classification using F1 scores. For APEACH, labels regarding hate and non-hate speech serve as positive and negative samples, respectively. In contrast, since BEEP! was initially formulated as a ternary task, we reformulate it into `hate'+`offensive' and `none' for the consistency of the binary setting. 

\subsection{Results}
The model-wise experimental results using BEEP! and APEACH are in  Table~\ref{table1}. It was encouraging that the models trained with BEEP! training set shows reasonable performance even in our dataset, which implies that our criteria for dataset generation are largely aligned with the existing work. 

\paragraph{Influence of corpus domain}
\label{sec:domain}
In the case of BEEP!, KcBERT-Large displays the highest performance, while in APEACH, KoELECTRA which generally scores lower in BEEP! shows almost the same performance as KcBERT. This implies that the style and domain of the dominant corpus used for pretraining of each PLM influence the downstream task performance. 



\begin{table}[ht]
\centering
\resizebox{0.6\columnwidth}{!}{%
\begin{tabular}{|c|c|}
\hline
\textbf{Topic}        & \textbf{F1 Score}        \\
\hline\hline
Nationality           & 0.8519 \\
Age and social status & 0.8700 \\
Eating habits         & 0.8182 \\
Appearance            & 0.8114 \\
Gender stereotypes     & 0.7993 \\
Sexual harassment     & 0.7610 \\
Racism                & 0.8511 \\
Origin and residence  & 0.8393 \\
Disabled              & 0.8525 \\
Education             & 0.9035 \\  
\hline
\end{tabular}%
}
\caption{\label{result-by-topic}
SoongsilBERT-Base's F1 score of binary classification according to topics. 
}
\label{table2}
\end{table}

\paragraph{Performance per topic} 
We observed the deviation of the inference accuracy by ten topics presented in the guideline (Table~\ref{table2}). This deviation seems to come from the difference between the construction scheme of the training corpus and the evaluation corpus. In detail, providing the random order of prompts while generating the hate speech data yielded diverse topics which are difficult to obtain unless the annotation corpora are collected from multiple web communities. We did not yet exactly find why F1 scores regarding gender stereotypes and sexual harassment are lower than other categories despite the dominance of gender-related instances in BEEP! corpus. One possibility is that the style of text regarding gender and sexuality differs much from that of BEEP!, yielding discrepancy between train and evaluation set. Ironically, this is one evidence of the domain coverage of our dataset not only regarding topics but also style, as to be investigated in the following section.

\section{Analysis}
\begin{figure}[hbt!]
   \centering
   \includegraphics[width=0.9\columnwidth]{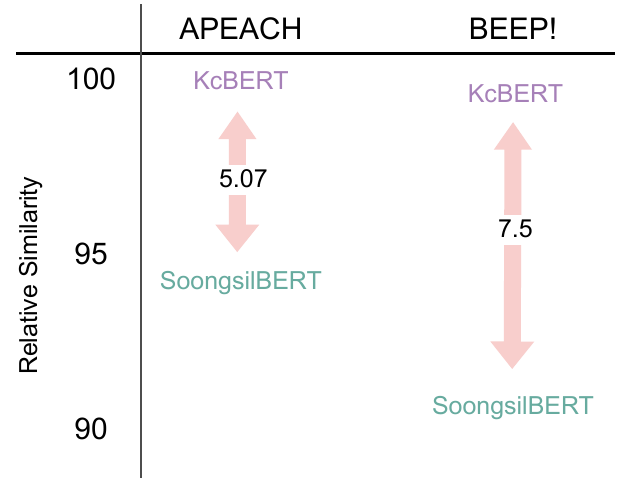}
   \caption{Averaged TF-IDF cosine similarity between the evaluation datasets and PLM training corpora. TF-IDF vectors are generated with the pretraining corpus of KcBERT and SoongsilBERT, and the dev set of BEEP! and APEACH.} \label{fig:figure7}
\end{figure}

\subsection{Domain Generalizability}
As discussed above, APEACH tackles domain dependency issues of annotation-based corpus construction scheme by i) letting the crowd generate the hate speech based on prompts and ii) not specifying the style of text that is created. These two are difficult to be guaranteed by annotating crawled web data from just several communities, and it is a current limitation of BEEP! which serves as a unique hate speech benchmark for Korean.
In detail, for BEEP! dev set where the text comes from the news comments domain, KoELECTRA, which lacks news comments in pretraining, shows relatively lower performance, while the tendency differs in APEACH.
This suggests that APEACH allows the investigation of the performance of fine-tuned hate speech detection models with less dependency on the domain of the corpus used for pretraining.
Besides, for SoongsilBERT
, a model trained with an augmented corpus upon the KcBERT's, 
the tendency for BEEP! and APEACH differs, implying that the domain-specificity of BEEP! dev set might have over-represented the advantage of KcBERT on news comment text.
In other words, the better performance of KcBERT in BEEP! shows that crawl and annotation-based dataset construction can bring dependency on specific domains, which acts as a limitation in the domain generalizability in the evaluation.

\paragraph{Quantification of domain generalizability}

To quantify these ideas, we validated the domain generalizability of each evaluation set by calculating the TF-IDF similarity between each set and the PLM pretraining corpus(Figure~\ref{fig:figure7}). Four scores were calculated using two pretraining corpora and two evaluation datasets, and normalized by the maximum value. Here, the TF-IDF word dictionary is built upon all whitespaced words of four corpus (APEACH, BEEP! dev set, KcBERT pretraining corpus, and SoongsilBERT pretraining corpus), and 1\% of KcBERT and SoongsilBERT pretraining corpus are randomly sampled for the feasibility of computing cosine similarity between the evaluation set instances. As a result, in the BEEP! dev set, the performance gap between KcBERT and SoongsilBERT is significant, whereas in APEACH, the gap is relatively less displayed. This implies that APEACH is less sensitive to the overlap with PLM pretraining corpus, suggesting the generalizability of the evaluation.

\paragraph{Qualitative analysis}

Among APEACH, we investigated the toxic examples regarding `Gender stereotypes' and `Sexual harassment', which were observed most significant among BEEP! train set. Two subtypes of texts were excerpted, namely the web comment style (that resembles BEEP!) and other formal texts. {\color{brown} WARNING: This part contains contents that may offend the readers.}

\begin{enumerate}
    \item ㅅㅂ옷 꼬라지 하고는.. 게이냐? (\textit{Wtf what a fashion... are you a gay?})
    \item 여자 연예인들은 다들 뒤로 스폰끼고 구멍 장사하는거 아닌가? (\textit{Isn't it a routine for idol girls to have a sugar daddy and fuck around?})
    \item 아무래도 아이를 돌보거나 그 밖의 집안일에 관해서는 남성들보다는 여성들이 좀 더 신경 써야 하는 것이 당연하다고 생각합니다. (\textit{I think it's natural for women to pay more attention to caring for their children or other household chores than men.})
    \item 남자면 그냥 해도 될텐데 고집을 많이 부리네요. 불편해도 좀 참아야 다들 편한데 (\textit{If you are a man, you can just do it; you have so much words. Even if it's uncomfortable, just endure it.})
\end{enumerate}

We checked that the SoongsilBERT-Base trained upon BEEP! train set correctly infers the toxicity of sentence (1-2) but fails to detect the harm of (3-4), where the stereotype is implicated in a polite and formal manner. This shows how the constructed dataset helps domain-generalized evaluation of hate speech detection, compared to the previous approach which adopts a single-domain text.



\subsection{Mitigating Train-Test Overlap}
Regarding contents, KcBERT was pretrained based on large-scale politics news comments. BEEP!  deals mainly with celebrity news comments, not politics, but shares a similar domain.
Therefore, a potential token overlap exists between KcBERT's pretraining corpus and BEEP!'s train/dev set, as we checked previously. This seems to boost the score of KcBERT significantly when evaluated with the BEEP! dev set.
We attempt to mitigate this in APEACH, which contains only crowd-generated thus unique utterances, preventing the over-representation of KcBERT shown in BEEP! dev set. 
APEACH guarantees such generalizability by generating text in a free-style manner based on topic prompts. Accordingly, we confirm that it fits with the evaluation of PLMs pretrained with the wider range of corpus (SoongsilBERT), compared to the BEEP! dev set.
Through this, we tackle again the risk of train-test overlap for the hate speech data constructed with crawling and annotation, and emphasize that APEACH mitigates this issue. This property also guarantees the utility of APEACH as a training set.
\paragraph{Training with APEACH}
For comprehensive understanding of the quality of APEACH as both training and evaluation set. We find-tuned two pretrained language models (KcBERT, KoELECTRA) with APEACH and BEEP! train set, and the evaluation results with APEACH and BEEP! dev set are shown in Table~\ref{tab:trainapeach}. We obtained similar results from both models in training with BEEP! and evaluating with APEACH, while difference of about 0.04 was displayed in evaluating with BEEP! dev set, which tells that BEEP! dev is more sensitive to pretraining corpora. However, we observed that the F1 score shifts down almost 0.04 lower for both models when the training is also done with APEACH. We first assumed that the size of training set (BEEP! train - 8K, APEACH - 3.7K) matters, and further conjecture that the different composition of BEEP! and APEACH influences.
\begin{table}[]
\centering
\resizebox{1.1\columnwidth}{!}{
\begin{tabular}{|c|cc|cc}
\cline{1-3}
                & \multicolumn{2}{c|}{\textbf{Validation}}                                                                                            &   &  \\ \cline{1-3}
\textbf{Train}           & \multicolumn{1}{c|}{\textbf{APEACH}} & \textbf{BEEP! dev}                                                               &  &  \\ \cline{1-3}
\textbf{APEACH} & \multicolumn{1}{c|}{-}                    & \begin{tabular}[c]{@{}c@{}}KoELECTRA: 0.7502\\ KcBERT-Large: 0.7893\end{tabular} &  &  \\ \cline{1-3}
\textbf{BEEP! train} &
  \multicolumn{1}{c|}{\begin{tabular}[c]{@{}c@{}}KoELECTRA: 0.8101\\ KcBERT-Large: 0.8116\end{tabular}} &
  \begin{tabular}[c]{@{}c@{}}KoELECTRA: 0.7916\\ KcBERT-Large: 0.8295\end{tabular} &
   &
   \\ \cline{1-3}
\end{tabular}
}

\caption{Evaluation results (F1 score) on APEACH and BEEP! with KoELECTRA and KcBERT-Large. Since APEACH does not have a specific training set, we exclude the case where the training set and the evaluation set are both APEACH.}
\label{tab:trainapeach}

\end{table}

\section{Conclusion}
In this work, we introduce a crowd-driven generation scheme in constructing an evaluation set for hate speech detection, distinct from the existing corpus construction schemes based on crawling and annotation. After a managed human text generation that ensures both the participants' anonymity and the reliability of the corpus, we report a thorough analysis of the created data, accompanied by a comparison with the prior work in Korean. The resulting corpus, APEACH, displays the potential of adopting crowd-driven generation in hate speech dataset construction, achieving generalizability and topic variety. Though there is headroom for the scalability of the corpus, we believe that the proposed scheme can be utilized to make up the evaluation set and training set of domain-agnostic hate speech detection. 

\section*{Ethical Consideration and Societal Implications}

Our study aims at the construction of hate speech corpus distinguished from the conventional scheme of crawling and annotation. This not only lessens the annotators' mental damage, which is probable in reading other people's toxic comments, but also mitigates the potential issue of license and privacy in distributing the corpus. First, we obtain the texts written by `workers' acknowledged by the moderator, not unknown `users', to make an appropriate compensation ($\approx$\$ 0.2 per sentence) and encourage the high-quality generation. Second, we guarantee anonymity but accept only qualified people, to prevent the case that the text is copy and pasted from other sources. Last, by ordering the omission of personally identifiable information in the generation process, we avoid the danger of information leaks in the final dataset. Our study is approved by institutional review board, where the specification is to be revealed after acceptance.

Our work incorporates several limitations and potential harms as well. First, using our scheme does not guarantee the hate speech dataset that satisfies everybody, since the intuition of workers differs significantly across various groups of people. Also, though our workers are selected after a pilot study, they may not be fully equipped with the ethical guideline. Thus, their decision might not always be ideal, which brings the degradation of reliability to the final label. Last, our binary scheme for hate speech lacks score-based decision and span notation which are up-to-date in the hate speech community, providing only hard-labeled instances of anonymous workers. However, we think the strength of our dataset is in initiating a generation-based hate speech detection corpus that allows crowd participation with lessened privacy and license concerns. We also want to state that the intended use of our dataset is to evaluate pretrained language models' ability to detect toxic language, with less dependency on the type of pretraining corpora, the domain of text, and the topic and length of sentences.

\section*{Acknowledgments}
The earliest version of this project, which was not
mentioned in this paper, was organized in 2018
in "YourSSU" and "SSUML", student research societies of Soongsil University. We would like to
thank Jongchan Kim and members of these two groups for giving inspiration to the kick-off
of the hate-speech project and the completion of
this manuscript.

We are extremly grateful to Moontae Lee, Jungwoo Ha, and
Seungwon Park for giving generous feedback and
suggestion to improve the quality of this paper. We
appreciate all the considerate reviews, including
anonymous ones, which enabled the development
of this research. Also, we thank DeepNatural AI, a
crowdsourcing platform, for serving as a moderator
of this project and taking care of worker management.

Finally, we give special thanks to Hyeongseok
Oh, Sungwon Lyu, and other Context Team members of Kakao Enterprise, for their utmost care and
attention in running this project and completing the
paper.

\bibliographystyle{acl_natbib}
\bibliography{anthology,acl2021}

\begin{thebibliography}{16}
\expandafter\ifx\csname natexlab\endcsname\relax\def\natexlab#1{#1}\fi

\bibitem[{Clark et~al.(2020)Clark, Luong, Le, and Manning}]{clark2020electra}
Kevin Clark, Minh-Thang Luong, Quoc~V. Le, and Christopher~D. Manning. 2020.
\newblock \href {https://openreview.net/pdf?id=r1xMH1BtvB} {{ELECTRA}:
  Pre-training text encoders as discriminators rather than generators}.
\newblock In \emph{ICLR}.

\bibitem[{Davidson et~al.(2017)Davidson, Warmsley, Macy, and
  Weber}]{hateoffensive}
Thomas Davidson, Dana Warmsley, Michael Macy, and Ingmar Weber. 2017.
\newblock Automated hate speech detection and the problem of offensive
  language.
\newblock In \emph{Proceedings of the 11th International AAAI Conference on Web
  and Social Media}, ICWSM '17, pages 512--515.

\bibitem[{Devlin et~al.(2019)Devlin, Chang, Lee, and
  Toutanova}]{devlin2019bert}
Jacob Devlin, Ming-Wei Chang, Kenton Lee, and Kristina Toutanova. 2019.
\newblock Bert: Pre-training of deep bidirectional transformers for language
  understanding.
\newblock In \emph{Proceedings of the 2019 Conference of the North American
  Chapter of the Association for Computational Linguistics: Human Language
  Technologies, Volume 1 (Long and Short Papers)}, pages 4171--4186.

\bibitem[{Hosseinmardi et~al.(2015)Hosseinmardi, Mattson, Rafiq, Han, Lv, and
  Mishra}]{hosseinmardi2015detection}
Homa Hosseinmardi, Sabrina~Arredondo Mattson, Rahat~Ibn Rafiq, Richard Han, Qin
  Lv, and Shivakant Mishra. 2015.
\newblock Detection of cyberbullying incidents on the instagram social network.
\newblock \emph{arXiv preprint arXiv:1503.03909}.

\bibitem[{Huang et~al.(2020)Huang, Xing, Dernoncourt, and
  Paul}]{huang-etal-2020-multilingual}
Xiaolei Huang, Linzi Xing, Franck Dernoncourt, and Michael~J. Paul. 2020.
\newblock \href {https://www.aclweb.org/anthology/2020.lrec-1.180}
  {Multilingual {T}witter corpus and baselines for evaluating demographic bias
  in hate speech recognition}.
\newblock In \emph{Proceedings of the 12th Language Resources and Evaluation
  Conference}, pages 1440--1448, Marseille, France. European Language Resources
  Association.

\bibitem[{Kiela et~al.(2020)Kiela, Firooz, Mohan, Goswami, Singh, Ringshia, and
  Testuggine}]{NEURIPS2020_1b84c4ce}
Douwe Kiela, Hamed Firooz, Aravind Mohan, Vedanuj Goswami, Amanpreet Singh,
  Pratik Ringshia, and Davide Testuggine. 2020.
\newblock \href
  {https://proceedings.neurips.cc/paper/2020/file/1b84c4cee2b8b3d823b30e2d604b1878-Paper.pdf}
  {The hateful memes challenge: Detecting hate speech in multimodal memes}.
\newblock In \emph{Advances in Neural Information Processing Systems},
  volume~33, pages 2611--2624. Curran Associates, Inc.

\bibitem[{National Institute~of Korean~Languages(2020)}]{nikl2020corpora}
NIKL National Institute~of Korean~Languages. 2020.
\newblock \href {https://corpus.korean.go.kr} {{NIKL CORPORA} 2020 (v.1.0)}.

\bibitem[{Lee(2020)}]{lee2020kcbert}
Junbum Lee. 2020.
\newblock Kcbert: Korean comments bert.
\newblock In \emph{Proceedings of the 32nd Annual Conference on Human and
  Cognitive Language Technology}, pages 437--440.

\bibitem[{Moon et~al.(2020)Moon, Cho, and Lee}]{moon-etal-2020-beep}
Jihyung Moon, Won~Ik Cho, and Junbum Lee. 2020.
\newblock \href {https://doi.org/10.18653/v1/2020.socialnlp-1.4} {{BEEP}!
  {K}orean corpus of online news comments for toxic speech detection}.
\newblock In \emph{Proceedings of the Eighth International Workshop on Natural
  Language Processing for Social Media}, pages 25--31, Online. Association for
  Computational Linguistics.

\bibitem[{Paszke et~al.(2019)Paszke, Gross, Massa, Lerer, Bradbury, Chanan,
  Killeen, Lin, Gimelshein, Antiga, Desmaison, Kopf, Yang, DeVito, Raison,
  Tejani, Chilamkurthy, Steiner, Fang, Bai, and
  Chintala}]{NEURIPS2019_bdbca288}
Adam Paszke, Sam Gross, Francisco Massa, Adam Lerer, James Bradbury, Gregory
  Chanan, Trevor Killeen, Zeming Lin, Natalia Gimelshein, Luca Antiga, Alban
  Desmaison, Andreas Kopf, Edward Yang, Zachary DeVito, Martin Raison, Alykhan
  Tejani, Sasank Chilamkurthy, Benoit Steiner, Lu~Fang, Junjie Bai, and Soumith
  Chintala. 2019.
\newblock \href
  {https://proceedings.neurips.cc/paper/2019/file/bdbca288fee7f92f2bfa9f7012727740-Paper.pdf}
  {Pytorch: An imperative style, high-performance deep learning library}.
\newblock In \emph{Advances in Neural Information Processing Systems},
  volume~32. Curran Associates, Inc.

\bibitem[{Sanh et~al.(2019)Sanh, Debut, Chaumond, and
  Wolf}]{sanh2019distilbert}
Victor Sanh, Lysandre Debut, Julien Chaumond, and Thomas Wolf. 2019.
\newblock Distilbert, a distilled version of bert: smaller, faster, cheaper and
  lighter.
\newblock \emph{arXiv preprint arXiv:1910.01108}.

\bibitem[{Stumpf et~al.(2007)Stumpf, Rajaram, Li, Burnett, Dietterich,
  Sullivan, Drummond, and Herlocker}]{stumpf2007toward}
Simone Stumpf, Vidya Rajaram, Lida Li, Margaret Burnett, Thomas Dietterich,
  Erin Sullivan, Russell Drummond, and Jonathan Herlocker. 2007.
\newblock Toward harnessing user feedback for machine learning.
\newblock In \emph{Proceedings of the 12th international conference on
  Intelligent user interfaces}, pages 82--91.

\bibitem[{Vaswani et~al.(2017)Vaswani, Shazeer, Parmar, Uszkoreit, Jones,
  Gomez, Kaiser, and Polosukhin}]{vaswani2017attention}
Ashish Vaswani, Noam Shazeer, Niki Parmar, Jakob Uszkoreit, Llion Jones,
  Aidan~N Gomez, {\L}ukasz Kaiser, and Illia Polosukhin. 2017.
\newblock Attention is all you need.
\newblock \emph{Advances in Neural Information Processing Systems},
  30:5998--6008.

\bibitem[{Waseem and Hovy(2016)}]{waseem2016hateful}
Zeerak Waseem and Dirk Hovy. 2016.
\newblock Hateful symbols or hateful people? predictive features for hate
  speech detection on twitter.
\newblock In \emph{Proceedings of the NAACL student research workshop}, pages
  88--93.

\bibitem[{Williams et~al.(2018)Williams, Nangia, and
  Bowman}]{williams2018broad}
Adina Williams, Nikita Nangia, and Samuel Bowman. 2018.
\newblock A broad-coverage challenge corpus for sentence understanding through
  inference.
\newblock In \emph{Proceedings of the 2018 Conference of the North American
  Chapter of the Association for Computational Linguistics: Human Language
  Technologies, Volume 1 (Long Papers)}, pages 1112--1122.

\bibitem[{Wolf et~al.(2020)Wolf, Debut, Sanh, Chaumond, Delangue, Moi, Cistac,
  Rault, Louf, Funtowicz, Davison, Shleifer, von Platen, Ma, Jernite, Plu, Xu,
  Le~Scao, Gugger, Drame, Lhoest, and Rush}]{wolf-etal-2020-transformers}
Thomas Wolf, Lysandre Debut, Victor Sanh, Julien Chaumond, Clement Delangue,
  Anthony Moi, Pierric Cistac, Tim Rault, Remi Louf, Morgan Funtowicz, Joe
  Davison, Sam Shleifer, Patrick von Platen, Clara Ma, Yacine Jernite, Julien
  Plu, Canwen Xu, Teven Le~Scao, Sylvain Gugger, Mariama Drame, Quentin Lhoest,
  and Alexander Rush. 2020.
\newblock \href {https://doi.org/10.18653/v1/2020.emnlp-demos.6} {Transformers:
  State-of-the-art natural language processing}.
\newblock In \emph{Proceedings of the 2020 Conference on Empirical Methods in
  Natural Language Processing: System Demonstrations}, pages 38--45, Online.
  Association for Computational Linguistics.

\end{thebibliography}


\clearpage

\appendix

\section{Pseudo Classifier}
\label{sec:supplementary}

\subsection{Dataset}
We use the texts of an online community (YourSSU) \footnote{Online community of a Korean university \url{https://yourssu.com/}} and news comments for the tagging of pseudo-gold labels. About 500,000 sentences were finally collected and were pre-processed with, e.g., removing duplication.
The labeling policy of the dataset partly follows \citet{hosseinmardi2015detection}.
We initially constructed a dictionary made up of about 200 profanity terms (toxic words), and labeled a text as "Hate" if it contains any of the listed terms. In specific, the toxic words were chosen referring to the criteria on politics/sexual/racism/religion, from the COC in PyCon KR. Profanity terms that are used as simple expressives, e.g., \texttt{fxxx}, as an interjection, were not included in the dictionary to prevent the false alarm.

\subsection{Model}
We deployed 
with the DistilKoBERT, \footnote{\url{https://github.com/monologg/DistilKoBERT}} namely the distilled version of KoBERT.
The performance of this model was measured with an F1 score of 0.5857 using APEACH.

\subsection{Demo Page}
\label{sec:appendix-demo}
The demo page shown in Figure~\ref{fig:figure2} was implemented using Flask, \footnote{\url{https://flask.palletsprojects.com/en/}} and was uploaded to the server using Heroku. \footnote{\url{https://www.heroku.com/}}
Due to the model size being too large to deploy on Heroku, we conducted quantization for linear operation and exported it using TorchScript. 

\section{Post-Labeling Information}
\label{sec:post-labeling}
During the post-labeling process of APEACH, the number of sentences accepted/rejected by the task-manager is summarized in Table~\ref{tab:postlabel}. The column `tutorial session' in Table~\ref{tab:postlabel} denotes the session that was conducted to guarantee the dataset quality (Section~\ref{sec:post_labeling}). The workers who have passed the tutorial session participated in the main session. APEACH is a dataset that includes both tutorial and main session.

\begin{table}[ht]
\centering
\resizebox{0.85\columnwidth}{!}{%
\begin{tabular}{|c|c|c|c|c|}
\hline
                \multicolumn{1}{|c}{}&\multicolumn{2}{|c|}{Tutorial session}&\multicolumn{2}{|c|}{Main session} \\
                \cline{2-5}
                & \textbf{Non-hate} & \textbf{Hate} & \textbf{Non-hate} & \textbf{Hate} \\
                \hline
\textbf{Accept} & 453               & 478           & 1386              & 1499          \\
\textbf{Reject} & 38                & 52            & 116               & 1             \\
\textbf{Total}  & 491               & 530           & 1502              & 1500  \\  
\hline
\end{tabular}%
}
\caption{The number of accepted/rejected sentences in the post-labeling process. The workers with more sincerity (less faults) in the tutorial finally participated in the main session. }
\label{tab:postlabel}
\end{table}

\section{Training Details}
\label{sec:training-detail}
Fine-tuning the pretrained language models bases on Huggingface Transformers \cite{wolf-etal-2020-transformers} 3.3.1 and PyTorch \cite{NEURIPS2019_bdbca288} 1.6.0. 
For all the pretrained models, checkpoints available in Huggingface Model Hub were used. \footnote{\url{https://huggingface.co/models}} GPU used for the fine-tuning is 1x NVIDIA TITAN RTX. We train the models for 10 epochs, with the batch size 32 and the learning rate of 5e-05.

\end{document}